# GPAFormer: Graph-guided Patch Aggregation Transformer for Efficient 3D Medical Image Segmentation

Chung-Ming Lo
Institute of Artificial Intelligence Innovation, Industry Academia Innovation School National Yang Ming Chiao Tung University
buddylo@nycu.edu.tw

I-Yun Liu
Department of Computer Science and Information Engineering National Chung Cheng University
a0958581118@gmail.com

Wei-Yang Lin
Department of Computer Science and Information Engineering, National Chung Cheng University
wylin@cs.ccu.edu.tw

*Abstract*—Deep learning has been widely applied to 3D medical image segmentation tasks. However, due to the diversity of imaging modalities, the high-dimensional nature of the data, and the heterogeneity of anatomical structures, achieving both segmentation accuracy and computational efficiency in multi-organ segmentation remains a challenge. This study proposed GPAFormer, a lightweight network architecture specifically designed for 3D medical image segmentation, emphasizing efficiency while keeping high accuracy. GPAFormer incorporated two core modules: the multi-scale attention-guided stacked aggregation (MASA) and the mutual-aware patch graph aggregator (MPGA). MASA utilized three parallel paths with different receptive fields, combined through planar aggregation, to enhance the network's capability in handling structures of varying sizes. MPGA employed a graph-guided approach to dynamically aggregate regions with similar feature distributions based on inter-patch feature similarity and spatial adjacency, thereby improving the discrimination of both internal and boundary structures of organs. Experiments were performed on public whole-body CT and MRI datasets including BTCV, Synapse, ACDC, and BraTS. Compared to the existed 3D segmentation networkd, GPAFormer using only 1.81 M parameters achieved overall highest DSC on BTCV (75.70%), Synapse (81.20%), ACDC (89.32%), and BraTS (82.74%). Using consumer level GPU, the inference time for one validation case of BTCV spent less than one second. The results demonstrated that GPAFormer balanced accuracy and efficiency in multi-organ, multi-modality 3D segmentation tasks across various clinical scenarios especially for resource-constrained and time-sensitive clinical environments.

## I. INTRODUCTION

Medical imaging technology has made substantial advancements since the late 20th century and have become essential part in modern healthcare. Modalities such as computed tomography (CT) and magnetic resonance imaging (MRI) not only improve diagnostic accuracy but also enhance the efficiency of clinical decision-making [1]. In the United States, the utilization rate of CT among adults increased from 56 per 1,000 person-years in 2000 to 141 per 1,000 person-years in 2016, whereas MRI utilization increased from 16 to 64 during the same period [2]. It is estimated that by 2055, the overall demand for medical imaging examinations may increase by approximately 16.9% to 26.9% compared with 2023 [3]. These findings highlight the growing reliance on medical imaging across various clinical environments.

3D medical imaging modalities such as CT and MRI provide more complete spatial information than 2D images to observe the overall position and morphology of organs. CT has high resolution and good ability to show tissue structures make it useful for accurately examining the anatomy and abnormalities of internal organs [4], and it is often used to observe abdominal organs, for example in cancer staging of the esophagus, liver, pancreas, and kidneys [5]. MRI, on the other hand, has strong ability to distinguish soft tissues, which allows clear visualization of structures such as the brain, heart, and muscles [6]. However, manual delineation of organs to accurately evaluate suspected abnormalities or lesions in 3D medical images is time-consuming and affect by inter- and intra-operator variability, which compromises diagnostic accuracy and treatment evaluation [7]. Automated image segmentation addresses these issues by reducing processing time, enhancing consistency and objectivity, and minimizing human error. Consequently, it provides substantial clinical benefits for large-scale data analysis, disease monitoring, treatment assessment, and surgical planning [8].

With the development of deep learning technology, end-to-end learning mechanisms can automatically learn important structural information from annotated data, overcoming the limitations of manual design [9, 10]. The disadvantages are the increasing of high computational and memory costs. In time-critical scenarios such as emergency care, intraoperative navigation, and intensive care, rapid and accurate CT/MRI segmentation is essential for lesion localization, organ annotation, and surgical planning [4-6]. Reducing parameters and computational complexity shortens inference time, enabling real-time applications. Lightweight segmentation networks further support edge deployment on scanners or surgical robots, minimizing latency and transmission risks while ensuring timely and reliable clinical decisions.

For this purpose, this study proposed graph-guided patch aggregation Transformer (GPAFormer), a lightweight segmentation framework designed for deployment on low-cost RTX GPUs offering a practical solution for real-world clinical environments. GPAFormer employed a three-stage Transformer-based network with two modules. Multi-receptive field feature extraction module can effectively interpret anatomical structures of varying scales, enhancing robustness to heterogeneous organ morphologies. The other graph-guided patch aggregation module was developed to adaptively merge semantically similar patches during Transformer encoding, enabling efficient representation learning. The integration of the two modules during training preserves segmentation accuracy while effectively reducing computational cost.

## II. RELATED WORK

When deep learning was first applied to image segmentation, it was mainly based on the architecture of convolutional neural networks (CNNs). CNNs, with their ability to capture local features and their hierarchical feature learning mechanism, gradually developed a series of effective segmentation strategies and promoted the wide application of medical image segmentation[11-14].

Ronneberger et al. proposed the U-Net [12], which brought a major breakthrough to image segmentation. U-Net adopted a symmetric encoder-decoder structure and introduced an innovative skip connection mechanism. U-Net was particularly suitable for medical image segmentation tasks where annotated data were scarce and boundary precision was critical. Building on U-Net, nnU-Net [14] was developed as a general-purpose self-configuring framework in medical image segmentation. nnU-Net aimed at addressing the challenge of poor generalization and heavy manual parameter tuning in deep learning networks when applied to different medical image datasets. An ensemble strategy, combining predictions from multiple network configurations was adopted. nnU-Net achieved Dice similarity coefficients (DSC) of 97.4% and 85.1% in kidney and kidney tumor segmentation, respectively [14]. Even so, the convolution operation with its local receptive field still has potential limitations in capturing long-range dependencies in images, making it difficult to effectively model global relationships between voxels or regions. In complex organ segmentation tasks that require global understanding, Transformer architectures, which can effectively learn long-range relationships, show more advantages [15].

Vision Transformer (ViT) [16] is derived from the Transformer architecture in natural language processing [15]. The input image is divided into multiple fixed-size, non-overlapping patches. Each patch is then flattened into a one-dimensional vector and linearly projected into a unified embedding space, forming the corresponding patch embeddings. Additionally, learnable positional embeddings and class token were added in the sequence of patch embeddings. By self-attention, the semantic information of the entire image was integrated. UNETR [17], building on ViT, and combined it with the U-Net architecture for 3D image segmentation. In the encoder, the multi-head self-attention mechanism can learn and build long-range dependencies across the entire 3D volume, helping to understand the complex 3D anatomical structures of organs, including overall shapes and relative positional relationships. UNETR achieved a DSC of 85.3% in BTCV abdominal organ segmentation [18]. nnFormer [19] interleaved convolution and self-attention in the backbone network, and to use local volume-based multi-head self-attention (LV-MSA) and global volume-based multi-head self-attention (GV-MSA) to build a feature pyramid and provide a sufficiently large receptive field. nnFormer achieved a DSC of 86.5% on the Synapse dataset for abdominal multi-organ segmentation [18].

Deep learning segmentation architectures in medical imaging have gradually introduced the Transformer into the encoder and combined it with convolutional layers to form hybrid architectures, which combine the local feature extraction of convolution with the global perception of the Transformer. The later developments focused on either accuracy or efficiency. Swin-UNETR [20], which focused on accuracy, adopted Swin Transformer as the encoder [21], and designed a hierarchical structure and a shifted window attention mechanism. The hierarchical structure built multi-level feature maps, which helped in capturing multi-scale features and provided receptive fields at different levels. The shifted window attention restricted self-attention computation within local windows and allowed information exchange across windows by shifting them, thereby enlarging the receptive field. Swin-UNETR achieved a DSC of 91.8% on BTCV abdominal organ segmentation [22]. However, Swin Transformer introduced high computational demands. SegFormer3D [23], which focused on efficiency, was the 3D version of SegFormer [24]. It adopted sequence reduction attention and a lightweight multi-layer perceptron decoder. The key and value sequences were reduced through linear projection, thereby lowering the spatial dimension of the key and value. This reduced the computational complexity from O(N2) to O(N2/R), where R is the reduction ratio. SegFormer3D contained only 4.5M parameters and achieved a DSC of 90.96% on the ACDC dataset [23]. UNETR++[25] also achieved efficiency in its design through efficient paired attention (EPA). It projected the key and value into a lower-dimensional space, so that the self-attention computation reached linear complexity with respect to the input sequence length. At the same time, it shared the weights of the query and key between spatial attention and channel attention, further reducing the number of parameters. UNETR++ achieved a DSC of 87.22% on the Synapse dataset for abdominal multi-organ segmentation [25].

Networks with excessive parameters have high learning capacity but are prone to overfitting when training data are limited, resulting in poor generalization to unseen cases. This issue is critical in medical imaging, where data scarcity and inter-institutional variability are common. For instance, Swin-UNETR, with its large parameter count, is unsuitable for clinical deployment due to the need for powerful hardware and its long inference time. Such oversized models limit real-time applications like diagnosis and intraoperative navigation, underscoring the necessity for lightweight and efficient architectures.

SegFormer3D [23] and UNETR++ [25] reduced the computational burden of attention by linearly shortening the key and value sequences. However, each sequence contained both important and less relevant components, and uniform linear reduction cannot efficiently preserve critical information such as organ boundaries. To address this limitation, this study proposed GPAFormer, which converted image structures into graphs and utilized both node similarity and adjacency relationships as the basis for attention sequence aggregation. This approach effectively shortened the sequence while improving both feature representation and computational efficiency achieving an optimal balance between segmentation

performance and efficiency. The GPAFormer include:

- Three-stage aggregation strategy: simplified the encoder and eliminating an entire downsampling level to achieve a substantial reduction in both parameter count and computational complexity, leading to accelerated inference speed. This architecture would be suitable for real-time applications or deployment on resource-constrained hardware.

- Multi-scale attention-guided stacked aggregation (MASA): constructed multiple feature extraction paths with heterogeneous receptive fields, enabling coordinated integration of multi-scale spatial semantics.

- Mutual-aware patch graph aggregator (MPGA): builded semantic relationships between patches using a graph structure, performing adaptive patch aggregation. The aggregation weights are dynamically adjusted according to semantic similarity and spatial boundaries.

## III. MATERIALS AND METHODS

### A. CT and MRI Datasets

The CT and MRI datasets used in the experiment for the segmentation evaluation included the publicly available MICCAI 2015 Multi-Atlas Abdomen Labeling Challenge (BTCV) dataset [18], the Synapse multi-organ segmentation (Synapse) dataset [18], the Automated Cardiac Diagnosis Challenge (ACDC) dataset [26], and the Multimodal Brain Tumor Segmentation Challenge (BraTS) dataset [27].

1) MICCAI 2015 Multi-Atlas Abdomen Labeling Challenge(BTCV) DATASET

The BTCV dataset consists of 50 portal venous phase abdominal CT scans collected under the supervision of institutional review board from colorectal cancer and hernia studies. Image volumes range from 512×512×85 to 512×512×198, with fields of view between 280×280×280 mm³ and 500×500×650 mm³, in-plane resolutions of 0.54–0.98 mm², and slice thicknesses of 2.5–5.0 mm. Thirty scans were annotated for 13 abdominal organs including the spleen, right kidney, left kidney, gallbladder, esophagus, liver, stomach, aorta, inferior vena cava, portal vein and splenic vein, pancreas, right adrenal gland, and left adrenal gland. by radiologists at Vanderbilt University Medical Center.

2) Synapse multi-organ segmentation (Synapse) Dataset

The Synapse dataset, a subset of BTCV, includes 30 portal venous phase CT scans with imaging characteristics consistent with BTCV. Synapse provides labels for 8 major organs: aorta, gallbladder, left kidney, right kidney, liver, pancreas, spleen, and stomach.

3) AUTOMATED CARDIAC DIAGNOSIS CHALLENGE (ACDC) DATASET

The ACDC dataset comprises cardiac MRIs of 150 patients collected at Dijon University Hospital over six years using 1.5T/3.0T scanners with steady-state free precession sequences under breath-hold. Images have an in-plane resolution of 1.34–1.68 mm²/pixel and 5 mm slice thickness. The end-diastolic (ED) and end-systolic (ES) images of 100 patients (200 cases) were manually annotated for the right ventricular cavity, myocardium, and left ventricular cavity by two experts following strict guidelines. Covering healthy and four pathological categories, ACDC is the largest publicly available, fully annotated cardiac MRI dataset.

4) MULTIMODAL BRAIN TUMOR SEGMENTATION CHALLENGE (BraTS) DATASET

The Medical Segmentation Decathlon (MSD) Task01_BrainTumour dataset is derived from the multi-institutional pre-operative adult glioma cohorts of BraTS 2016–2017 and was released following standardized preprocessing—co-registration to a common anatomical template, resampling to 1 mm³ isotropic resolution, and skull-stripping. In total, it comprises 750 four-dimensional (4D) volumes, of which 484 provide voxel-wise manual annotations for training and 266 are unlabeled. Each case includes four structural MRI sequences: T1-weighted (T1w), contrast-enhanced T1-weighted (T1Gd/T1ce), T2-weighted (T2w), and FLAIR. Annotations strictly follow the BraTS tumor subregion definitions, encompassing peritumoral edema (ED), non-enhancing/necrotic tumor core (NCR/NET), and enhancing tumor (ET), in addition to background. For evaluation, labels are commonly aggregated into the nested regions whole tumor (WT), tumor core (TC), and enhancing tumor (ET).

## B. GPAFormer

Deep learning segmentation networks based on Transformer architectures were capable of modeling long-range dependencies corresponding to complete 3D anatomical spatial information. But the computational cost of the self-attention increased quadratically with the input sequence length. To address this issue, GPAFormer (Fig. 1) was developed as a lightweight architecture achieving generalization across imaging modalities (CT and MRI). GPAFormer had a three-stage network architecture, with the number of blocks in each stage being (2, 2, 2). Through enhanced multi-receptive-field feature extraction and a dynamic patch aggregation network, the conventional four-stage structure was reduced to three stages. Feature maps of various dimensions from encoding were integrated by a lightweight multi-layer perceptron (MLP) to achieve efficient and high-quality segmentation. During encoding, MASA was capable of extracting comprehensive receptive field features with relevant weights. The following MPGA which adaptively guided the self-attention through dynamic graph aggregation, merging redundant adjacent patches into larger units. This reduced sequence length while focusing on organ boundaries with rich information.

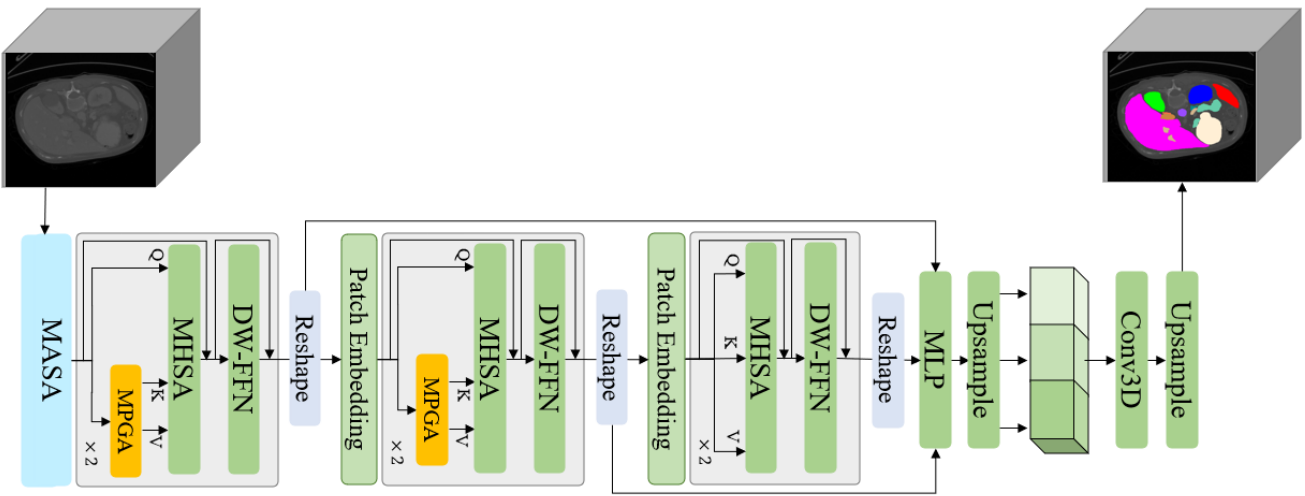

Fig. 1. The architecture of GPAFormer

1) multi-scale attention-guided stacked aggregation module (MASA)

MASA (Fig. 2) adopted three independent and parallel feature extraction paths. After obtaining spatial structural information from diverse convolution kernels of different scales, the features were fused through planar stacking and then enhanced with a self-attention mechanism, allowing more effective integration of contextual information from different receptive field regions and cross-regional feature relationships.

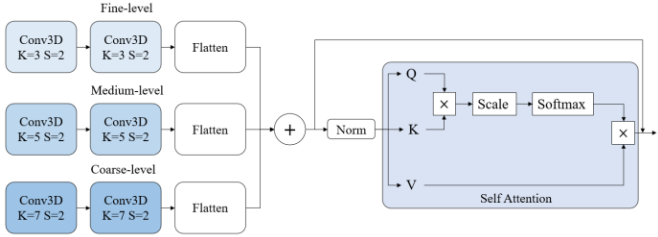

Fig. 2. Multi-scale attention-guided stacked aggregation

The three feature extraction paths consisted of fine-level, medium-level, and coarse-level paths, implemented and interpreted as follows:

- Fine-level path: Used a smaller convolution kernel (3×3×3), where the smaller receptive field captured boundary contours, texture details, and high-frequency features. This enhanced the ability to resolve small structures, thin tissues, and organ edges in the image, helping to ensure accurate segmentation boundaries and separation between organs.

- Medium-level path: Used a medium-sized convolution kernel (5×5×5), achieving a balance between receptive field coverage and detail preservation. It can capture medium-scale morphological features while retaining a certain level of structural detail. In multi-organ segmentation tasks, it helped to analyze morphological differences and relative spatial relationships between adjacent organs, and strengthened the continuity between coarse- and fine-level features.

- Coarse-level path: Used a larger convolution kernel (7×7×7) to cover a wider spatial range at once, capturing large-scale morphological features of organs. It effectively depicted overall structural contours, positional relationships, and morphological trends in the image, which is helpful for identifying large anatomical regions or low-frequency structural patterns.

When generating feature maps with different receptive fields, each path applied resolution reduction with stride=2, performed twice in sequence, reducing the resolution to one-half and one-quarter of the original. After completing the downsampling in each path, the feature representations extracted at different receptive fields were flattened into vectors and then aggregated through element-wise summation, as shown in Equation

(4):

$$F_{fused} = F_F + F_M + F_C \qquad (4)$$

where $F_{fused}$ is the aggregated feature map, and $F_F$, $F_M$, and $F_C$ are the vectorized feature representations from the coarse-, medium-, and fine-level paths, respectively.

By aggregating the fine-, medium-, and coarse-level feature maps through element-wise summation, both micro details and macro structures can be obtained simultaneously. This feature complementarity reduced the limitations of a single receptive field. The coarse-level feature map was responsible for locating the overall structural range, while the medium- and fine-level feature maps gradually supplemented local details and boundary information. This design effectively addressed the recognition of adjacent multi-organ and morphological variations, while also resolving differences in contrast, texture, and noise between CT and MRI, thereby improving cross-modality generalization.

In the later part of the MASA module, the aggregated feature map $F_{fused}$, which contained both multi-granularity features and anatomical positional information, was fed into a self-attention layer to capture deeper structural relationships and global context within the feature map. At the same time, a residual connection was applied to facilitate information flow and maintain feature integrity. The final output of the MASA module is given in Equation (5):

$$F_{MASA} = F_{fused} + SelfAttention(F_{fused}) \qquad (5)$$

The overall design of MASA focused on integrating multi-level spatial information and regional features under different receptive fields. Through multiple feature extraction paths in a parallel architecture, MASA can capture discriminative spatial cues without adding extra network depth. Since multi-granularity convolution features were essentially still based on local receptive fields, their ability to model long-range dependencies was limited. By introducing the self-attention mechanism, cross-regional feature relationships can be established on a global scale, compensating for the shortcomings of convolutional networks in long-range dependency modeling. This also strengthened the analysis of interactions among multiple organs, and enables adaptive capture of global context under different imaging conditions, improving adaptability across devices and scanning protocols.

2) mutual-aware patch graph aggregator module (MPGA)

MPGA, as shown in Fig. 3, used a graph structure to represent the topology of spatial adjacency relationships in the image, where each image patch is represented as a node and edges connect neighboring patches, forming a mutual-aware patch graph. This graph structure can explicitly represent the spatial topology between adjacent regions and implicitly capture semantic correlations across patches. Both the feature attributes of patch nodes and their spatial neighborhood were jointly used as similarity conditions to guide the adaptive patch aggregation mechanism.

The aggregation weights were adjusted based on semantic similarity (e.g., consistency within the same organ tissue) and spatial boundary constraints (e.g., differences at organ boundaries). Through a learnable assignment process, multiple similar patches were merged into representative object nodes. This dynamic aggregation reduced the sequence length, thereby reduced the computational load of attention, while simultaneously directing the network's representational capacity toward clinically relevant regions, especially organ boundaries and areas where structural interactions occur.

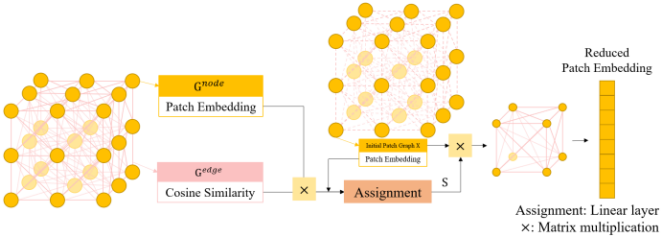

Fig. 3. Mutual-aware patch graph aggregator

The initial patch graph X is built by first converting each patch into a Transformer-encoded embedding vector (the large cubic with virtual edge in Fig. 3). For an image consisting of N patches, the embedding vectors are denoted as $\{p_i \in \mathbb{R}^d\}_{i=1}^N$ with corresponding 3D center coordinates $\{x_i \in \mathbb{R}^3\}_{i=1}^N$. An undirected weighted graph $X = (V, E)$ is then constructed. The edges are undirected, meaning the relationship between nodes A and B is symmetric, and edges are associated with values representing the strength or distance of their relationship. V denotes the set of nodes $\{p_i\}_{i=1}^N$, and E denotes the set of edges. If the center distance between two nodes $p_i, p_j$ satisfies $\|x_i - x_j\|_2 < \tau$, an undirected edge $(i, j) \in E$ is created. In the experiments, $\tau = 1.8$, covering a $3 \times 3 \times 3$ cubic neighborhood, so each patch can be connected to at most 26 neighboring patches.

The following steps describe the process of constructing the mutual-aware patch graph G (the large cubic with solid edge in Fig. 3) by building the node feature matrix and spatial adjacency matrix and calculating edge weights:

1. Node feature matrix $G^{node} \in \mathbb{R}^{N \times d}$: all patch embedding vectors are used as node features to form matrix $G^{node}$. The i-th row corresponds to the embedding vector $p_i \in \mathbb{R}^d$ of the i-th patch, i.e., $G^{node} = [p_1; p_2; \ldots; p_N]$

2. Spatial adjacency matrix $A^{spatial} \in \mathbb{R}^{N \times N}$: built based on the edge set described above, reflecting whether two patches are spatial neighbors.

3. Similarity matrix $A^{sim} \in \mathbb{R}^{N \times N}$: during message passing, each node must retain its own feature, thus $A_{ii}^{sim} = 1$. For each edge $(i, j)$, the edge weight is defined by the cosine similarity between patch embedding vectors, as shown in Equation (6):
$$A_{ij}^{sim} = \cos(\theta_{ij}) = \frac{p_i \cdot p_j}{\|p_i\| \cdot \|p_j\|} \qquad (6)$$

In Fig. 4, the element-wise multiplication of the spatial adjacency matrix $A^{spatial}$ and the similarity matrix $A^{sim}$ yields the similarity-weighted adjacency matrix $G^{edge} \in \mathbb{R}^{N \times N}$. The completed mutual-aware patch graph is then composed of the node feature matrix $G^{node}$ and the similarity-weighted adjacency matrix $G^{edge}$, which together reflect both spatial adjacency relationships and feature similarity between patches.

| | $p_1$ | $p_2$ | $\ldots$ | $p_N$ |
|---|---|---|---|---|
| $p_1$ | 1 | 1 | $\ldots$ | 0 |
| $p_2$ | 1 | 1 | $\ldots$ | 0 |
| $\vdots$ | $\vdots$ | $\vdots$ | $\ddots$ | $\vdots$ |
| $p_N$ | 0 | 0 | $\ldots$ | 1 |

$A^{spatial}$

$\odot$

| | $p_1$ | $p_2$ | $\ldots$ | $p_N$ |
|---|---|---|---|---|
| $p_1$ | 1 | 0.95 | $\ldots$ | 0.64 |
| $p_2$ | 0.95 | 1 | $\ldots$ | 0.54 |
| $\vdots$ | $\vdots$ | $\vdots$ | $\ddots$ | $\vdots$ |
| $p_N$ | 0.64 | 0.54 | $\ldots$ | 1 |

$A^{sim}$

$=$

| | $p_1$ | $p_2$ | $\ldots$ | $p_N$ |
|---|---|---|---|---|
| $p_1$ | 1 | 0.95 | $\ldots$ | 0 |
| $p_2$ | 0.95 | 1 | $\ldots$ | 0 |
| $\vdots$ | $\vdots$ | $\vdots$ | $\ddots$ | $\vdots$ |
| $p_N$ | 0 | 0 | $\ldots$ | 1 |

$G^{edge}$

Fig. 4. The construction of the similarity-weighted adjacency matrix

Next, each patch is enhanced with contextual information from its neighboring regions, and the enhanced information is combined with the initial patch graph X, as shown in Equations (7) and (8):
$$\tilde{X} = G^{edge} G^{node} + X \qquad (7)$$
$$\tilde{x_i} = \sum_{j=1}^N G_{ij}^{edge} G_j^{node} + x_i, \quad i = 1, 2, \ldots, N \qquad (8)$$

where $\tilde{x_i}$ represents the updated feature vector of the $i$-th node, while $x_i$ and $G_j^{node}$ denote the feature vectors of the $i$-th node in the initial patch graph and the $j$-th node in the mutual-aware patch graph, respectively. $G_{ij}^{edge}$ indicates the similarity weight between nodes $i$ and $j$. Through this design, each patch representation not only retains its own features but also incorporates a weighted sum of neighboring and similar nodes, further strengthening contextual relationships and feature consistency.

Finally, a learnable assignment process is applied to aggregate the enhanced local graph nodes, as shown in Equation (9):

$$S = \text{softmax}(\tilde{X}W_s) \in \mathbb{R}^{N \times K} \quad (9)$$

where $W_s \in \mathbb{R}^{d \times K}$ is a trainable parameter and K is the number of representative nodes after aggregation. The assignment matrix S indicates the degree to which each original patch belongs to the K representative nodes. Furthermore, the initial patch graph X is aggregated into a reduced representation using a weighted approach, as shown in Equation (10):

$$F_{MPGA} = S^T X \quad (10)$$

where $F_{MPGA} \in \mathbb{R}^{K \times d}$ is the aggregated node representation obtained by combining the initial patch graph X with the assignment matrix S, and $S^T$ denotes the transpose of the assignment matrix.

Compared with traditional graph neural network (GNN) architectures [28] such as GCN [29], where the adjacency matrix is defined by binary connectivity (1 if an edge exists, otherwise 0) and node features are updated through multi-layer message passing, MPGA strengthens contextual information with a one-time feature update after graph construction, avoiding the computational overhead and over-smoothing problems associated with multi-layer propagation. Moreover, the adjacency in MPGA is determined by cosine similarity between node features, rather than binary edge presence alone. Finally, a learnable assignment strategy is introduced to complete the sequence reduction.

## C. Evaluation Metrics

In the experiments, the DSC, which is widely used in medical image segmentation tasks, was adopted as the evaluation metric [30] to effectively quantify the degree of overlap between the segmented regions and the ground truth. The DSC is defined as follows:

$$DSC(O_{gt}, O_{pred}) = \frac{2|O_{gt} \bigcap O_{pred}|}{|O_{gt}| + |O_{pred}|} \quad (11)$$

where $O_{gt}$ represents the ground truth and $O_{pred}$ denotes the set of pixels in the region segmented by the model. The DSC ranges between 0 and 1, with higher values indicating greater consistency between the model's segmentation and the actual annotations, reflecting better segmentation performance. When the predicted segmentation perfectly matches the ground truth (i.e., $O_{gt} = O_{pred}$, the DSC reaches 1, representing a perfect segmentation.

The advantage of using DSC as an evaluation metric lies in its robustness to the class imbalance problem commonly seen in medical images (for example, when a certain organ occupies only a small portion of the image). Compared to metrics that focus solely on pixel-level accuracy, the DSC provides a more balanced consideration of false positives and false negatives, thereby offering a more comprehensive and accurate reflection of model performance in complex medical image segmentation tasks.

## IV. RESULT

In the training and validation of the four medical imaging datasets, a 5-fold cross-validation was adopted. Each dataset was evenly divided into five mutually exclusive subsets. In each training and validation cycle, four subsets were used as the training set, and the remaining one subset was used as the validation set.

The experiments were conducted on a workstation equipped with an Intel® Core™ i7-12700 CPU @ 4.90 GHz, 64 GB RAM, and an NVIDIA GeForce RTX 4090 GPU with 24 GB memory. The segmentation networks were implemented using PyTorch v1.12.1+cu113 [31] and the MONAI library [32].

For training, 3D sub-volumes of size 96×96×96 were randomly cropped from the BTCV, Synapse, and BraTS datasets, while 96×96×16 sub-volumes were cropped from the ACDC dataset. The sub-volumes were sampled with a 1:1 ratio between foreground and background. Afterward, the sub-volumes were divided into fixed-size image patches within the network, with the initial patch size set to 4×4×4.

The data augmentation strategies included both spatial and intensity transformations. For the spatial aspect, random flipping along the x, y, or z axes (RandFlipd) and random 90-degree rotations (RandRotate90d) were applied. For intensity variation, random shifting of voxel intensities (RandShiftIntensityd) was employed.

The AdamW optimizer [33] was used which prevents overfitting by decoupling weight decay from gradient computation. Weight decay, also known as L2 regularization, constrains model complexity by adding a penalty term to the loss function. The formula is defined as Eq. (12):

$$L_{2\_reg} = \lambda \sum_{i=1}^{N} \theta_i^2 \quad (12)$$

where $\lambda$ is the regularization coefficient, controlling the weight of the penalty term; $\theta_i$ denotes the $i$-th parameter of the model; and N is the total number of model parameters. Compared with the traditional Adam optimizer, AdamW applies weight decay independently after the parameter update step, enabling more effective regularization.

The batch size was set to 1, and the learning rate was set to 0.0001. An early stopping mechanism was employed: if the performance on the validation set did not improve for 10 consecutive validation steps, training was automatically stopped. Each validation was performed every 500 iterations.

The loss function used was DiceCE loss, which combines Dice loss and cross-entropy (CE) loss. Dice loss measures the spatial overlap between the segmentation results and the ground truth:

$$\text{Dice loss} = 1 - \frac{2\sum O_{gt} \cdot O_{pred}}{\sum O_{gt} + \sum O_{pred}} \tag{13}$$

where $O_{gt}$ denotes the pixel set of the ground truth region and $O_{pred}$ denotes the pixel set of the predicted region.

Cross-entropy loss measures the difference between the predicted and ground truth probability distributions:

$$\text{CE loss} = -\sum_{i=1}^{I} O_{gt,i} \log(O_{pred,i}) \tag{14}$$

where $I$ represents the number of regions, $O_{gt,i}$ is the probability of the $i$-th region in the ground truth, and $O_{pred,i}$ is the predicted probability of the $i$-th region.

Finally, the two losses were combined with weights α and β as in Eq. (15):

$$\text{DiceCEloss} = \alpha \cdot \text{Dice loss} + \beta \cdot \text{CE loss} \tag{15}$$

### A. Segmentation Accuracy

Compared with other aggregation methods such as strided convolution [34], linear projection [16], and pooling operations [35], which usually achieve efficiency by fixed regions or regular downsampling, GPAFormer adopted a similarity-guided dynamic aggregation strategy. This approach adaptively adjusted the aggregation process based on the feature distribution, so that patch compression was no longer dependent only on fixed strides or positional relationships, but instead groups patches according to local structural similarity. As shown in Fig. 5, the MPGA module introduced in GPAFormer (top row) can effectively cluster organ boundaries and internal regions into different groups when processing areas such as the liver and right kidney. In the liver region, the outer edge is aggregated into continuous and consistent patches, presenting a clear structural contour, while the interior consists mainly of large, unified aggregated patches, reflecting the model's good sensitivity to structural details of the liver. For the right kidney region, the aggregation results also generally align with its structural outline, with concentrated patches and relatively clear boundaries. In contrast, fixed aggregation methods (bottom row) often divide a single structure into multiple discontinuous aggregated patches in the above organ regions. Both the edges and interiors appear fragmented and scattered, making it difficult for the aggregation results to accurately correspond to the true anatomical structures, thus showing poorer overall spatial consistency.

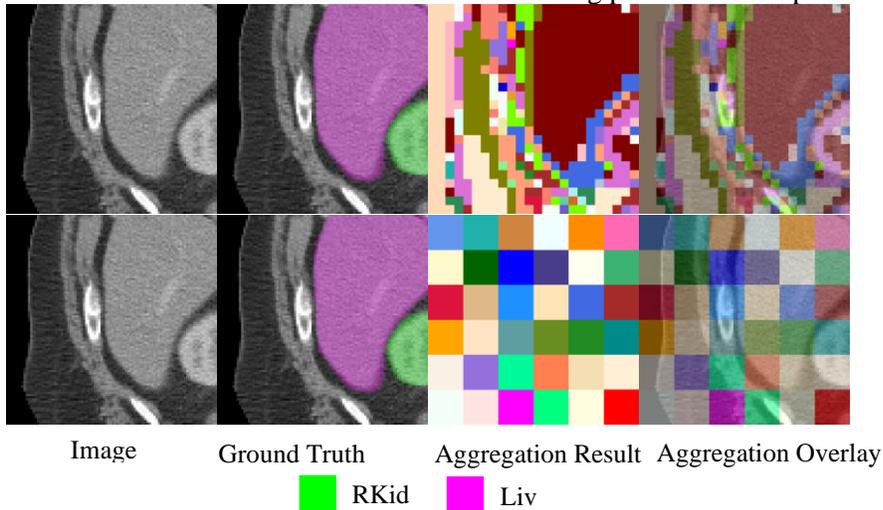

Image    Ground Truth    Aggregation Result    Aggregation Overlay

■ RKid    ■ Liv

Fig. 5. Visualization comparison between the MPGA module introduced in GPAFormer (top row) and fixed aggregation methods (bottom row).

In DSC comparisons, GPAFormer was compared with six 3D medical image segmentation networks, including nnUNet [14], nnFormer [19], UNETR [17], UNETR++ [25], SegFormer3D [23], and Swin-UNETR [20]. As shown in Table I, GPAFormer achieved an average DSC of 75.70% on the BTCV dataset, which was the highest among all evaluated segmentation networks. In addition, GPAFormer exhibited the lowest number of parameters, requiring only 40% of the parameters of the second-lowest model, SegFormer3D. Its FLOPs ranked among the top two lowest, and its inference time was within the top three fastest models. Table II shows that GPAFormer achieved the highest accuracies on most organs, including spleen (89.72%), right kidney (85.72%), left kidney (84.81%), liver (93.76%), stomach (82.29%), and aorta (88.81%).

Fig. 6 presents a visual comparison of segmentation results on the validation set of the BTCV dataset across different networks. From the visualization, it can be observed that GPAFormer produces boundaries and shapes that are closer to the ground truth for most organs, particularly in the segmentation of the stomach (Sto), where other networks often misclassify surrounding structures as the stomach, whereas GPAFormer is able to accurately delineate its contour. In addition, nnUNet and UNETR show clear omissions in the left adrenal gland (LAG), while UNETR and SegFormer3D exhibit over-segmentation in the liver (Liv) that extends into non-liver regions. Furthermore, nnUNet, nnFormer, UNETR, and UNETR++ demonstrate slight spill-over in the segmentation of the spleen (Spl).

TABLE I Performance comparison of different networks on the BTCV dataset.

| Network | DSC (%)↑ | Parameter (M)↓ | FLOPs (GFLOPs)↓ | Inf. Time (S) ↓ |
|---|---|---|---|---|
| nnUNet | 73.79 | 30.71 | 1297.62 | 18.1043 |
| nnFormer | 73.45 | 149.32 | 273.41 | 2.3627 |
| UNETR | 69.82 | 92.78 | 82.70 | 0.4817 |
| UNETR++ | 73.77 | 29.87 | 58.33 | 1.1782 |
| SegFormer3D | 70.32 | 4.50 | **5.01** | **0.1644** |
| Swin-UNETR | 75.05 | 62.19 | 329.20 | 2.2061 |
| GPAFormer | **75.70** | **1.81** | 24.05 | 0.9928 |

DSC=Dice Similarity Coefficient , Inf. Time= Inference  Time

TABLE II The DSC (%) results of BTCV dataset

| Network | Spl | RKid | LKid | Gal | Eso | Liv | Sto | Aor | IVC | PSV | Pan | RAG | LAG | DSC ↑ |
|---|---|---|---|---|---|---|---|---|---|---|---|---|---|---|
| nnUNet | 85.68 | 77.69 | 76.69 | 52.14 | 69.30 | 91.87 | 77.71 | 86.05 | **81.73** | 69.06 | 69.38 | 61.08 | 54.75 | 73.31 |
| nnFormer | 83.19 | 82.01 | 80.65 | **65.25** | 69.89 | 93.19 | 81.13 | 85.92 | 77.60 | 61.57 | 63.86 | 56.37 | 54.21 | 73.45 |
| UNETR | 84.70 | 83.06 | 80.48 | 53.84 | 66.55 | 92.26 | 72.97 | 83.49 | 75.72 | 57.19 | 54.56 | 55.04 | 47.74 | 69.82 |
| UNETR++ | 88.06 | 84.43 | 83.04 | 61.17 | 70.68 | 93.58 | 80.66 | 86.57 | 79.85 | 57.94 | 65.11 | 55.54 | 52.37 | 73.77 |
| SegFormer3D | 85.39 | 83.42 | 82.00 | 55.22 | 62.70 | 93.10 | 75.21 | 84.79 | 76.88 | 57.79 | 61.16 | 48.56 | 47.97 | 70.32 |
| Swin-UNETR | 87.21 | 82.80 | 80.70 | 63.58 | **71.89** | 93.24 | 78.76 | 87.68 | 81.15 | 63.59 | 65.72 | **61.28** | **57.99** | 75.05 |
| GPAFormer | **89.72** | **85.72** | **84.81** | 63.53 | 71.23 | **93.76** | **82.29** | **88.81** | 80.95 | 63.88 | 69.15 | 53.57 | 56.65 | **75.70** |

spleen (Spl), right kidney (RKid), left kidney (LKid), gallbladder (Gal), esophagus (Eso), liver (Liv), stomach (Sto), aorta (Aor), inferior vena cava (IVC), portal and splenic vein (PSV), pancreas (Pan), right adrenal gland (RAG), and left adrenal gland (LAG)

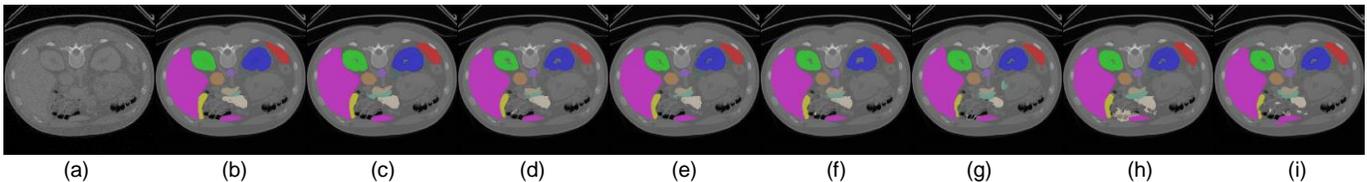

Spl RKid LKid Gal Eso Liv Sto Aor IVC PSV Pan RAG LAG

Fig. 6. Segmentation results on BTCV dataset (a) the original CT image (b) ground truth (c) GPAFormer (Ours) (d) nnUNet (e) nnFormer (f) Swin-UNETR (g) UNETR (h) UNETR++ (i) SegFormer3D

With respect to the Synapse dataset with less labels (Table III), GPAFormer achieved an average DSC of 81.20%, outperforming all compared networks. In addition, GPAFormer demonstrated accuracy comparable to or even surpassing that of higher-parameter networks across most organs. Fig. 7 presents a visual comparison of segmentation results on the validation set of the Synapse dataset across different networks. GPAFormer generally produces contours that are closer to the ground truth, with stable performance in boundary continuity and shape adherence, particularly for organs such as the liver (Liv) and left kidney (LKid). In contrast, nnUNet, nnFormer, and SegFormer3D incorrectly segment part of the liver (Liv) region as the pancreas (Pan), while SegFormer3D misclassifies the pancreas (Pan) as the stomach (Sto). Additionally, nnUNet, nnFormer, Swin-UNETR, and UNETR exhibit segmentation results with holes in the left kidney (LKid). Moreover, nnFormer underestimates the pancreas region, yielding a shape that deviates considerably from the annotation.

TABLE III The DSC (%) results of Synapse dataset

| Network | Aor | Gal | LKid | RKid | Liv | Pan | Spl | Sto | DSC↑ |
|---|---|---|---|---|---|---|---|---|---|
| nnUNet | 82.66 | 55.21 | 75.10 | 80.34 | 92.85 | 68.11 | **89.43** | 79.81 | 77.94 |
| nnFormer | 85.81 | 59.83 | 82.87 | 85.11 | **93.39** | 65.34 | 87.46 | 78.57 | 79.80 |
| UNETR | 83.62 | 57.28 | 80.46 | 82.45 | 92.32 | 54.24 | 84.47 | 74.47 | 76.16 |
| UNETR++ | **87.92** | 62.64 | 83.77 | 85.91 | 93.38 | 65.44 | 87.37 | 81.14 | 80.95 |
| SegFormer3D | 85.43 | 57.75 | 84.34 | 84.45 | 92.79 | 60.27 | 86.74 | 76.38 | 78.52 |
| Swin-UNETR | 87.68 | 64.68 | 80.92 | 83.85 | 93.35 | 66.81 | 85.93 | 75.73 | 79.87 |
| GPAFormer | 87.02 | **69.28** | **87.75** | **86.15** | 91.92 | **69.50** | 75.46 | **82.48** | **81.20** |

aorta (Aor), gallbladder (Gal), left kidney (LKid), right kidney (RKid), liver (Liv), pancreas (Pan), spleen (Spl), stomach (Sto)

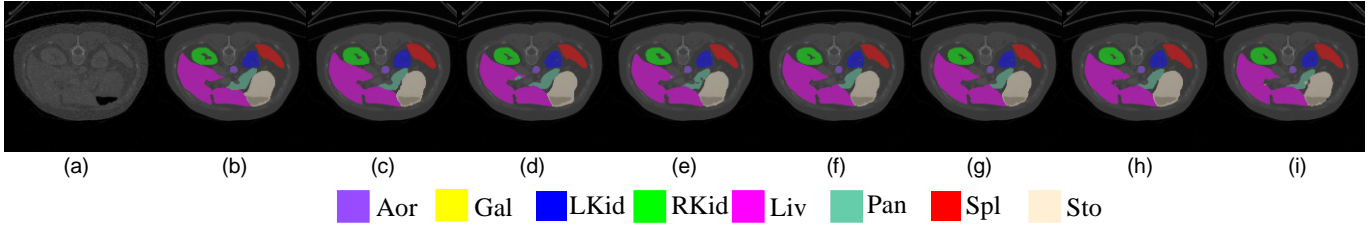

(a)  (b)  (c)  (d)  (e)  (f)  (g)  (h)  (i)

Aor  Gal  LKid  RKid  Liv  Pan  Spl  Sto

Fig. 7. Segmentation results on Synapse dataset (a) the original CT image (b) ground truth (c) GPAFormer (Ours) (d) nnUNet (e) nnFormer (f) Swin-UNETR (g) UNETR (h) UNETR++ (i) SegFormer3D

Table IV shows that GPAFormer achieves an average DSC of 89.32% on the ACDC dataset, representing the best performance among all compared networks. For individual structures, GPAFormer attains DSCs of 87.92% for the right ventricle (RV), 86.68% for the myocardium (Myo), and 93.35% for the left ventricle (LV), ranking first or second across all networks. Fig. 8 presents a visual comparison of segmentation results on the validation set of the ACDC dataset. The figure shows that GPAFormer produces contours closer to the ground truth for RV, Myo, and LV, with particularly strong performance in distinguishing between Myo and LV. In contrast, UNETR, UNETR++, and SegFormer3D often predict Myo regions that expand beyond the annular area, while UNETR++ also exhibits discontinuities along the LV boundary. For RV, all methods show varying degrees of under-segmentation; however, GPAFormer is able to preserve a more complete shape and boundary coverage, resulting in outputs that are comparatively closer to the ground truth.

TABLE IV The DSC (%) results of ACDC dataset

| Network | RV | Myo | LV | DSC↑ |
|---|---|---|---|---|
| nnUNet | 86.18 | 85.27 | 92.49 | 87.98 |
| nnFormer | 86.81 | 85.95 | 93.05 | 88.60 |
| UNETR | 80.47 | 81.55 | 90.18 | 84.07 |
| UNETR++ | 87.29 | 86.63 | **93.41** | 89.11 |
| SegFormer3D | 80.65 | 78.09 | 89.24 | 82.66 |
| Swin-UNETR | 85.77 | **87.47** | 92.99 | 88.74 |
| GPAFormer | **87.92** | 86.68 | 93.35 | **89.32** |

TABLE V The DSC(%) result of BraTS dataset

| Network | WT | ET | TC | DSC↑ |
|---|---|---|---|---|
| nnUNet | 88.53 | 79.16 | 74.61 | 80.76 |
| nnFormer | 87.44 | 73.17 | 78.24 | 79.61 |
| UNETR | 78.90 | 58.50 | 76.10 | 71.10 |
| UNETR++ | 88.89 | 78.99 | 75.27 | 81.05 |
| SegFormer3D | 88.73 | 73.41 | 80.10 | 80.74 |
| Swin-UNETR | 89.71 | **76.97** | 81.30 | 82.66 |
| GPAFormer | **89.92** | 76.03 | **82.27** | **82.74** |

right ventricle (RV), myocardium (Myo), left ventricle (LV)          whole tumor (WT), enhancing tumor (ET), tumor core (TC)

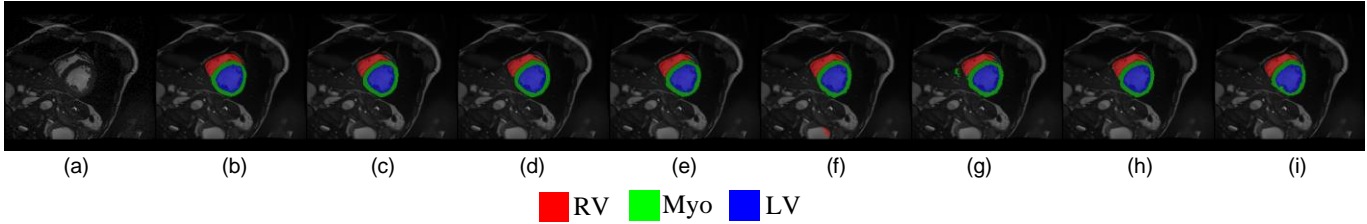

RV ■ Myo ■ LV ■

Fig. 8. Segmentation result on ACDC dataset (a) the original MRI image (b) ground truth (c) GPAFormer (Ours) (d) nnUNet (e) nnFormer (f) Swin-UNETR (g) UNETR (h) UNETR++ (i) SegFormer3D

Regarding the comparison on the BraTS dataset in Table V, GPAFormer obtained the overall DSC=82.74%, which was the highest among all compared networks. In terms of sub-regions, GPAFormer obtained a better overall trade-off among WT, ET, and TC. Consequently, its average DSC ranked first. This was consistent with the observed boundary stability and shape conformity in the figures. Fig. 9 compared the segmentation results of various networks on the BraTS validation set. From the figure, it was observed that GPAFormer has boundaries and shapes for the edema (ED), the non-enhancing tumor (NCR/NET), and the enhancing tumor (ET) more closely matched the ground truth annotations. This was obvious at the interface between ET and NCR/NET, where a cleaner boundary was maintained with fewer misclassifications. In comparison, UNETR and UNETR++ occasionally showed discontinuous or expanded boundaries. nnUNet and nnFormer sometimes showed local over-segmentation and small holes in the tumor core region. SegFormer3D's edema edge was at times blurry, and while the overall boundary of Swin-UNETR was smooth, it slightly underestimated the quantification in the ET. Overall, GPAFormer demonstrated a more stable performance in terms of boundary continuity.

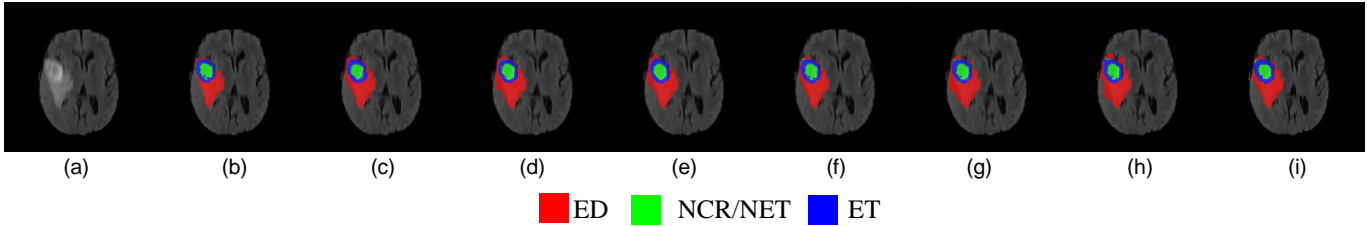

ED ■ NCR/NET ■ ET ■

Fig. 9. Segmentation result on BraTS dataset (a) the original MRI image (b) ground truth (c) GPAFormer (Ours) (d) nnUNet (e) nnFormer (f) Swin-UNETR (g) UNETR (h) UNETR++ (i) SegFormer3D

### B. Ablation Experiment

To further verify the contribution of the two key modules of GPAFormer to the overall performance, an ablation study was performed to analyze the effectiveness of MASA and MPGA. Table VI shows the average DSC of each configuration on the BTCV dataset. As a comparison baseline, the average DSC without any modules was 71.22%. When only the MPGA module was introduced, the average DSC increased to 72.54%. When only the MASA module was added, the average DSC reached 73.17%. When both modules were applied simultaneously in the GPAFormer architecture, the average DSC improved from 71.22% to 75.70%, achieving 6.3% improvement.

TABLE VI Ablation results on the BTCV dataset measured by DSC

| Method | MASA | MPGA | DSC↑ |
|---|---|---|---|
| Baseline | ✗ | ✗ | 71.22 |
| + MPGA | ✗ | ✓ | 72.54 |
| + MASA | ✓ | ✗ | 73.17 |
| GPAFormer | ✓ | ✓ | **75.70** |

### V. Discussion

In medical image analysis, building an end-to-end neural network capable of automatically segmenting organs in 3D CT and MRI images brings substantial clinical value. The automatic segmentation process can reduce errors and inconsistencies caused by manual operations in an objective and efficient manner [7, 8]. However, most segmentation networks in the current literature are not designed for real-time or off-line clinical demands [14, 17]. Their large number of parameters and computational complexity often require reliance on cloud servers [19, 20]. In high time-critical settings such as intraoperative navigation or emergency care, network latency can compromise surgical safety [36]. Hospitals also express concerns about data security and privacy risks when uploading medical images to the cloud [36]. Furthermore, small and medium-sized hospitals or resource-limited regions cannot afford the cost of extensive computational resources. A lightweight network architecture, capable of running directly on scanners or edge devices, avoids the latency, network dependency, and security risks of cloud computation, ensuring stable usability in urgent, intraoperative, and intensive care scenarios.

To address this, this study proposed GPAFormer, a lightweight architecture designed to maintain sufficient segmentation accuracy. Within its staged segmentation framework, GPAFormer introduced two key modules: MASA and MPGA. MASA aggregated convolutional kernels of different receptive fields through multiple stacked paths, then utilized attention mechanisms to assign optimized sequence weights, thereby covering diverse spatial ranges during feature extraction and strengthening correlations among fine-, medium-, and coarse-scale structures. Based on these aggregated features, MPGA constructed a graph structure from the spatial relationships among patches and employed inter-patch similarity as a learnable soft label to guide aggregation. This enabled patches that are spatially adjacent and semantically similar to merge into representative nodes, while maintaining independence for heterogeneous patches. Such dynamic graph-based aggregation directed the attention mechanism toward organ boundaries and inter-organ contours, while substantially reducing redundant internal comparisons. In contrast, traditional static sequence reduction strategies, such as stride convolution [34], linear projection [16], and pooling [35], tend to indiscriminately merge highly different patches, especially at organ boundaries, often causing fragmented morphology and limiting segmentation quality.

As shown in Table I, GPAFormer had the fewest parameters (1.81 M) among all compared segmentation networks, while nnFormer had the largest (149.32 M). GPAFormer also achieved the second-lowest computational cost (24.05 GFLOPs). In terms of segmentation accuracy across datasets, GPAFormer achieved the highest DSC scores to support both CT (BTCV, Synapse) and MRI (ACDC, BraTS) image dataset. and performed stably across different organ labels. Visual comparisons further showed that GPAFormer can more accurately delineate complete contours of the liver, stomach, and kidneys, without leakage or gaps at boundaries. In particular, for organs with clearly defined raw boundaries, GPAFormer produced smoother contours, highlighting the MPGA module's ability to direct attention toward edge definition.

In terms of inference efficiency, GPAFormer achieved an average inference time of 0.9928 seconds per case on BTCV, making it the most accurate among all models with inference times under one second. By comparison, Swin-UNETR required 2.2061 seconds per inference. This highlighted GPAFormer's superior balance between accuracy and efficiency, making it particularly suitable for critical clinical applications [37, 38].

The ablation experiment further showed that combing MASA and MPGA yielded the best performance at 74.38%. MASA contributed diverse and enriched features, while MPGA provided structured aggregation, resulting in a "broaden then refine" effect that preserved both critical contours and global relationships while avoiding unnecessary internal comparisons, thus outperforming either module alone. GPAFormer's efficiency resulted from the coordinated behavior of its modules rather than any single component. MASA expanded the receptive field through sequential convolution and progressive downsampling, achieving effective feature integration with minimal overhead. Likewise, although ViT-style attention captured long-range dependencies, its cost was prohibitive in 3D settings. Existing static sequence-reduction methods reduced computation but at the expense of inprecise boundaries and structures. MPGA employed similarity-guided dynamic aggregation to compress patches while preserving key anatomical details, enabling an enlarged receptive field with a favorable balance between precision and efficiency.

GPAFormer demonstrated a strong balance between accuracy and efficiency, making it a practical candidate for clinical deployment. Nevertheless, further validation will be needed. The overall dataset size remained limited, and small-organ samples were underrepresented, which may hinder learning for small structures. For such cases, contrastive learning [39] can be applied to enhance discriminative features. Contrastive learning improved generalization in small-sample or class-imbalance settings by constructing positive and negative sample pairs, enabling the network to better distinguish similar but different structures and reduce issues with unclear boundaries. Self-supervised learning [40] pretraining can also be utilized to extract richer feature representations from unlabeled data, reducing reliance on annotated datasets and providing stronger initial performance in downstream segmentation tasks. Validation on local clinical data will be also necessary to assess generalization across diverse sources. In addition, more pathological cases should be incorporated to verify the network's adaptability in various conditions. In this situation, another potential improvement would be incorporating medical prior anatomical knowledge as broader spatial constraints such as using image-text alignment to align semantic vectors such as organ names or descriptions with image features, allowing the network to establish correspondence between anatomical semantics and image regions [41]. Addressing these areas will further strengthen GPAFormer's reliability and real-world utility.

## VI. Conclusion

GPAFormer was proposed with the goal of improving both the efficiency and accuracy of 3D medical image segmentation. The architecture integrated the MASA and MPGA modules to fuse multi-scale spatial information and perform patch aggregation through a learnable assignment strategy, respectively. Across the whole-body image datasets including BTCV (CT), Synapse (CT), ACDC (MRI), and BraTS (MRI), GPAFormer having the fewest parameters achieved the overall highest DSC compared to existed 3D segmentation networks. These experimental results indicated that GPAFormer was well-suited for applications in resource-constrained environments or scenarios requiring rapid inference.

## VII. Acknowledge

THIS WORK WAS SUPPORTED IN PART BY MINISTRY OF SCIENCE AND TECHNOLOGY IN TAIWAN (MOST 112-2221-E-A49-188-MY3)